\documentclass{article}

\usepackage{microtype}
\usepackage{graphicx}
\usepackage{subcaption}
\usepackage{booktabs} 
\usepackage{caption} 

\usepackage{multirow}
\usepackage[export]{adjustbox} 

\usepackage{algorithm}
\usepackage{hyperref}

\usepackage[preprint]{icml2026}

\usepackage{amsmath}
\usepackage{amssymb}
\usepackage{mathtools}
\usepackage{amsthm}

\usepackage[capitalize,noabbrev]{cleveref}

\theoremstyle{plain}

\theoremstyle{definition}

\theoremstyle{remark}

\usepackage[textsize=tiny]{todonotes}

\icmltitlerunning{Training-Free Cross-Architecture Merging for Graph Neural Networks}

\begin{document}
\icmlsetsymbol{equal}{*}

\twocolumn[
  \icmltitle{Training-Free Cross-Architecture Merging for Graph Neural Networks}

  \begin{icmlauthorlist}
    \icmlauthor{Rishabh Bhattacharya}{equal,iiit}
    \icmlauthor{Vikaskumar Kalsariya}{equal,iiit}
    \icmlauthor{Naresh Manwani}{iiit}
  \end{icmlauthorlist}

  \icmlaffiliation{iiit}{International Institute of Information Technology, Hyderabad, India}

  \icmlcorrespondingauthor{Rishabh Bhattacharya}{rishabh.bhattacharya@research.iiit.ac.in}
  \icmlcorrespondingauthor{Vikaskumar Kalsariya}{vikaskumar.kalsariya@research.iiit.ac.in}

  \icmlkeywords{Machine Learning, ICML}

  \vskip 0.3in
]

\printAffiliationsAndNotice{\icmlEqualContribution}
\begin{abstract}
Model merging has emerged as a powerful paradigm for combining the capabilities of distinct expert models without the high computational cost of retraining, yet current methods are fundamentally constrained to \textbf{homogeneous architectures}. For GNNs, however, message passing is \textbf{topology-dependent} and sensitive to misalignment, making direct parameter-space merging unreliable. To bridge this gap, we introduce \textbf{H-GRAMA} (Heterogeneous Graph Routing and Message Alignment), a training-free framework that lifts merging from \emph{parameter space} to \emph{operator space}. We formalize \textbf{Universal Message Passing Mixture (UMPM)}, a shared operator family that expresses heterogeneous GNN layers in a common functional language. H-GRAMA enables \textbf{cross-architecture GNN merging} (e.g., GCN$\leftrightarrow$GAT) without retraining, retaining high specialist accuracy in most cases in compatible depth settings and achieving inference speedups of $1.2\times$--$1.9\times$ over ensembles.
\end{abstract}

\section{Introduction}

The graph domain is beginning to surface its own merging-specific obstacles. Deep Graph Mating frames learning-free reuse for GNNs and highlights the sensitivity of interpolation to topology-dependent effects, proposing message coordination and message-statistics calibration as key ingredients \citep{jing2024deep}. More recently, \citet{garg2025gnnmerge} benchmark model merging for GNNs and propose embedding alignment to improve practical performance. These works substantiate that GNN merging is \emph{not} a direct transplant of vision/NLP merging recipes, and that controlling the \emph{message distribution} is central. However, existing formulations typically assume homogeneous architectures or address a different reuse goal than our \emph{specialist node-partition} setting: two parents trained on different labeled node subsets of the same graph, merged into a single child with no labels and no optimization at merge time.

We study a concrete and practically motivated composition problem in \emph{single-graph transductive node classification}. We are provided two pretrained \emph{specialist} parent GNNs trained under \emph{different labeled node subsets} (training masks) on the \emph{same} graph. Following the split-data protocol of \citet{ainsworth2023git}, we partition the training nodes into two disjoint subsets using a fixed \emph{specialist class split} (used only during parent training and evaluation): subset~A contains 80\% of labeled training nodes from one class group and 20\% from the other, while subset~B uses the complementary allocation. Parent $f_A$ is trained on subset~A and Parent $f_B$ on subset~B, while both models predict over the same label space $\{0,\dots,K-1\}$. The parents may be \emph{heterogeneous} (e.g., GCN \citep{kipf2017gcn}, GAT \citep{velickovic2018gat}, GraphSAGE \citep{hamilton2017graphsage}, GIN \citep{xu2019powerful}), and may differ in depth and width. Our goal is to construct a \emph{single} child model $f_C$ that preserves each parent's performance on its own specialist node subset, while running as one network at inference time and \emph{without} any gradient-based training, distillation, or fine-tuning during the merge.

At first glance, this objective resembles distillation, mixtures-of-experts, continual learning, or federated aggregation, but these paradigms typically require additional optimization, multi-expert inference, or architectural compatibility. Our setting instead requires a \emph{single} merged GNN produced \emph{training-free} and \emph{label-free} at merge time, while handling heterogeneous GNN operators and depth/width mismatch.

Recent work on \emph{training-free model merging} has shown that weight-space operations can be effective when models are sufficiently aligned. Stochastic Weight Averaging averages points along an SGD trajectory and improves generalization \citep{izmailov2018averaging}, and ``model soups'' average independently fine-tuned checkpoints that share a pretrained initialization \citep{wortsman2022model}. However, in the specialist regime and with heterogeneous GNN backbones, naive parameter mixing is brittle: the same function can be represented by many parameterizations due to permutation symmetries and representational misalignment. Alignment-aware methods address this in Euclidean domains by explicitly matching or permuting units, e.g., matched averaging in federated settings \citep{wang2020federated} and permutation-based ``teleportation'' such as Git Re-Basin \citep{ainsworth2023git}. Moreover, recent advances extend training-free merging to \emph{disjoint tasks} (typically with multi-head outcomes) \citep{stoica2024zipit} and propose dual-space constraints for pretrained model merging \citep{xu2024trainingfree}. While these results motivate training-free composition, they do not resolve the additional structure of graph message passing and the operator-level discrepancies across GNN architectures.

\paragraph{Our approach: merging in operator space.}
We propose Universal Message Passing Mixture \textsc{UMPM-Merge}, which lifts the merge problem from ``weights'' to a shared \emph{operator} language. Concretely, we express diverse GNN layers as mixtures over a fixed basis of message-passing operators (e.g., self-term, normalized GCN aggregation, sum/mean aggregation, etc.), providing a common representational envelope across architectures \citep{gilmer2017mpnn}. This enables \emph{transport and fusion} of heterogeneous layers after aligning their internal coordinates.

A second obstacle is that heterogeneous backbones do not share layer depths, widths, or activation parameterizations. We therefore align internal computations using representation similarity tools. We use linear Centered Kernel Alignment (CKA) \citep{kornblith2019similarity} to identify transport-safe correspondences between pre-activation representations, and then compute orthogonal (or semi-orthogonal) rectangular Procrustes maps \citep{schonemann1966generalized} to move parameters into a shared coordinate system. This design is consistent with the broader view that network representations can be related by simple linear transforms when architectures (or parts of them) are functionally comparable \citep{lenc2015understanding,bansal2021revisiting}.

Finally, even after alignment and operator-level fusion, we empirically observe systematic distribution shift in \emph{edge messages} induced by mixing operators from two parents, an issue also emphasized in recent learning-free GNN reuse literature \citep{jing2024deep}. \textsc{UMPM-Merge} therefore includes a deterministic \emph{message calibration} stage that matches low-order moments of aggregated edge-message statistics between parents and the child, implemented in an aggregate-space (folded) form that avoids per-edge materialization.

\paragraph{Contributions}
\begin{itemize}
  \item We formalize \emph{specialist merging} for \emph{single-graph transductive node classification}: merging two pretrained, potentially heterogeneous GNNs trained on disjoint labeled \emph{node subsets} of the same graph into one child model, with no labeled data and no gradient-based optimization at merge time.
  \item We introduce a \emph{universal message passing mixture} (UMPM) operator family that covers common GNN layer families within a shared operator vocabulary, enabling parameter transport and architecture-agnostic fusion.
  \item We propose a label-free, closed-form merge pipeline that aligns heterogeneous depths via CKA \citep{kornblith2019similarity}, transports parameters via Procrustes maps \citep{schonemann1966generalized}, fuses operators by closed-form gate regression and convex mixing, and stabilizes the child via scalable message-statistics calibration, building on the emerging understanding that message distributions are a primary failure mode in GNN merging \citep{jing2024deep,garg2025gnnmerge}.
\end{itemize}

\section{Related Work}

Graph Neural Networks (GNNs) are a standard tool for learning over relational data, with strong performance in node classification and related graph mining tasks \citep{kipf2017gcn,velickovic2018gat}. Modern GNN architectures can be viewed through the lens of message passing \citep{gilmer2017mpnn}, instantiated in popular variants such as GraphSAGE \citep{hamilton2017graphsage}, GIN \citep{xu2019powerful}, and attention-based models \citep{velickovic2018gat}. In many deployments, however, the practical bottleneck is not training a single GNN, but \emph{reusing and composing} multiple pretrained models that were produced under incompatible constraints (e.g., different architectures, different training policies, or different label availability).

\paragraph{Training-free model merging and alignment-aware methods.}
Early training-free methods rely on identical architectures and simple weight interpolation. Stochastic weight averaging (SWA) averages checkpoints along an SGD path \citep{izmailov2018averaging}, model soups extend this to multiple fine-tuned checkpoints \citep{wortsman2022soups}, and periodic SWA introduces scheduled averaging \citep{guo2022pswa}. Fisher-weighted averaging weighs models by their Fisher information \citep{matena2022fisher}, while RegMean uses linear regression without training data \citep{jin2023regmean}. Task arithmetic \citep{ilharco2023taskvec}, TIES-Merging \citep{yadav2023ties}, DARE \citep{yu2024dare}, and PCB-Merging \citep{du2024pcb} further refine parameter selection and scaling. To handle misaligned units, alignment-aware methods permute neurons before averaging: Git Re-Basin aligns hidden units to achieve barrier-free connectivity \citep{ainsworth2023git}, building on permutation invariance insights \citep{entezari2022permutation}. FedMA matches channels in federated learning \citep{wang2020fedma}, OT-Fusion uses optimal transport \citep{singh2020otfusion}, Sinkhorn-based rebasining learns differentiable transport plans \citep{pena2023sinkhorn}, and MuDSC jointly optimizes in weight and activation spaces \citep{xu2024muds}. These methods assume common operator blocks and Euclidean domains.

\paragraph{Merging under task and architectural heterogeneity.}
When specialists share no labels, ZipIt! merges via feature-level zipping \citep{stoica2024zipit}, AdaMerging learns layer-wise coefficients by minimizing entropy \citep{yang2024adamerging}, and UQ-Merge uses uncertainty estimates to guide merging order \citep{qu2024uqmerge}. For depth/width mismatches, MuDSC adapts group-norm and multi-head attention \citep{xu2024muds}, while training-free heterogeneous model merging uses representation similarity and elastic neuron zipping \citep{xu2024heteromerge}. These approaches demonstrate merging across vision and NLP models but do not consider graph-specific operators.

\paragraph{Graph-centric model reuse and GNN merging.}
GNNs pose unique challenges because message passing depends on topology. DistillGCN transfers structural knowledge via a local structure-preserving module \citep{yang2020distillgcn}, and AmalgamateGNN trains a slimmable student with topological attribution \citep{jing2021amalgamate}; both require training at merge time. Deep Graph Mating (GRAMA) formulates the first training-free GNN reuse task using dual-message coordination and message-statistics calibration \citep{jing2024deep}, while GNNMerge introduces task-agnostic node-embedding alignment with closed-form solutions \citep{garg2025gnnmerge}. Our proposed \textsc{H-GRAMA} extends this line by merging \emph{heterogeneous} GNNs without labels or training via a universal operator basis, CKA-Procrustes alignment, closed-form fusion, and message calibration.

\begin{figure*}[t]
  \centering
  \includegraphics[width=\textwidth]{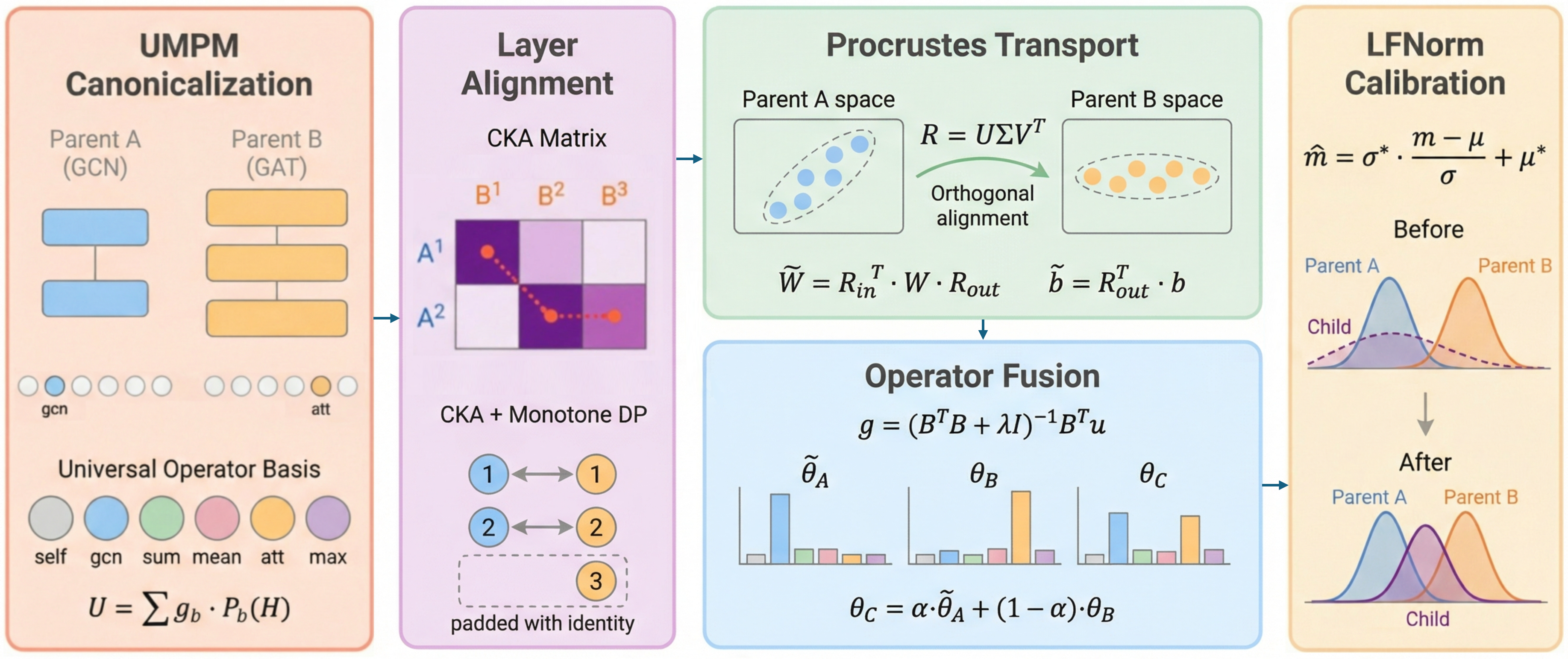}
\caption{\textbf{H-GRAMA pipeline.} Heterogeneous GNN parents are canonicalized to a universal operator basis (UMPM), aligned via CKA and Procrustes transport, fused through closed-form gate regression with confidence-weighted mixing ($\alpha$), and stabilized via LFNorm moment calibration.}
  \label{fig:myfig}
\end{figure*}

\begin{table*}[t]
\vskip 0.1in
\caption{Retention performance across datasets under varying architecture pairs, depths, and widths. Values show minimum retention $\min(\text{Ret}_A, \text{Ret}_B)$ averaged over 5 seeds. \textbf{Bold}: $\geq 0.80$. \underline{Underline}: $\geq 0.70$. All merges are training-free and label-free.}
\label{tab:main_results}
\vskip 0.1in
\begin{center}
\scshape
\resizebox{\textwidth}{!}{%
\begin{tabular}{@{}llccccccccccccccc@{}}
\toprule
& & \multicolumn{3}{c}{\textbf{Cora}} & \multicolumn{3}{c}{\textbf{CiteSeer}} & \multicolumn{3}{c}{\textbf{Actor}} & \multicolumn{3}{c}{\textbf{Amazon-Ratings}} & \multicolumn{3}{c}{\textbf{Arxiv}} \\
\cmidrule(lr){3-5} \cmidrule(lr){6-8} \cmidrule(lr){9-11} \cmidrule(lr){12-14} \cmidrule(lr){15-17}
Pair & Depth & 64-64 & 64-128 & 128-64 & 64-64 & 64-128 & 128-64 & 64-64 & 64-128 & 128-64 & 64-64 & 64-128 & 128-64 & 64-64 & 64-128 & 128-64 \\
\midrule
GCN--SAGE & 2-2 & \textbf{.80$\pm$.18} & \textbf{.80$\pm$.17} & \textbf{.85$\pm$.15} & \underline{.74$\pm$.25} & \underline{.72$\pm$.22} & .69$\pm$.23 & \textbf{.93$\pm$.06} & \textbf{.89$\pm$.08} & \textbf{.95$\pm$.05} & \textbf{.82$\pm$.22} & \textbf{.83$\pm$.15} & \textbf{.93$\pm$.06} & \textbf{.80$\pm$.14} & \textbf{.88$\pm$.06} & \underline{.71$\pm$.24} \\
& 2-3 & \textbf{.83$\pm$.20} & .52$\pm$.27 & \textbf{.90$\pm$.05} & \textbf{.87$\pm$.07} & \underline{.76$\pm$.14} & \textbf{.82$\pm$.15} & \textbf{.87$\pm$.08} & \textbf{.97$\pm$.02} & \textbf{.87$\pm$.11} & \underline{.73$\pm$.32} & \textbf{.89$\pm$.14} & \textbf{.90$\pm$.06} & .67$\pm$.17 & \textbf{.80$\pm$.10} & .67$\pm$.07 \\
& 3-2 & \underline{.78$\pm$.23} & \textbf{.85$\pm$.13} & .65$\pm$.20 & .63$\pm$.26 & \underline{.75$\pm$.20} & \underline{.75$\pm$.18} & \underline{.76$\pm$.12} & \underline{.71$\pm$.18} & \underline{.73$\pm$.12} & \textbf{.97$\pm$.06} & \textbf{1.00$\pm$.01} & \textbf{.94$\pm$.11} & \underline{.75$\pm$.20} & \textbf{.82$\pm$.11} & \textbf{.86$\pm$.11} \\
\midrule
GCN--GAT & 2-2 & \textbf{.89$\pm$.17} & \textbf{.95$\pm$.03} & \textbf{.88$\pm$.12} & \underline{.73$\pm$.15} & \textbf{.90$\pm$.12} & \underline{.75$\pm$.32} & \textbf{.96$\pm$.03} & \textbf{.95$\pm$.05} & \textbf{.95$\pm$.04} & \textbf{.92$\pm$.11} & \textbf{.91$\pm$.14} & \textbf{.95$\pm$.04} & \textbf{.93$\pm$.01} & \textbf{.90$\pm$.02} & \textbf{.92$\pm$.02} \\
& 2-3 & \textbf{.94$\pm$.06} & \underline{.77$\pm$.11} & \textbf{.94$\pm$.03} & \textbf{.85$\pm$.16} & \textbf{.88$\pm$.04} & \textbf{.86$\pm$.11} & \textbf{.96$\pm$.03} & \textbf{.95$\pm$.06} & \textbf{.80$\pm$.10} & \textbf{.90$\pm$.11} & \textbf{.92$\pm$.06} & \textbf{.96$\pm$.05} & \textbf{.81$\pm$.09} & \textbf{.84$\pm$.06} & \textbf{.81$\pm$.09} \\
& 3-2 & .68$\pm$.26 & \textbf{.81$\pm$.15} & .63$\pm$.29 & .54$\pm$.28 & \textbf{.85$\pm$.11} & \underline{.73$\pm$.15} & \textbf{.83$\pm$.16} & \textbf{.86$\pm$.11} & \textbf{.93$\pm$.06} & \textbf{.96$\pm$.09} & \textbf{1.00$\pm$.01} & \textbf{.93$\pm$.10} & \textbf{.82$\pm$.05} & .69$\pm$.30 & \textbf{.88$\pm$.05} \\
\midrule
SAGE--GAT & 2-2 & \textbf{.86$\pm$.12} & \textbf{.94$\pm$.06} & \textbf{.86$\pm$.09} & \underline{.70$\pm$.10} & \textbf{.87$\pm$.14} & \underline{.73$\pm$.13} & \textbf{.97$\pm$.02} & \textbf{.95$\pm$.02} & \textbf{.92$\pm$.04} & \textbf{.80$\pm$.14} & \textbf{.82$\pm$.13} & \textbf{.89$\pm$.05} & \textbf{.92$\pm$.01} & \textbf{.91$\pm$.01} & \textbf{.92$\pm$.03} \\
& 2-3 & \textbf{.90$\pm$.08} & \textbf{.94$\pm$.04} & \textbf{.88$\pm$.12} & \textbf{.94$\pm$.03} & \textbf{.95$\pm$.03} & \textbf{.82$\pm$.13} & \textbf{.81$\pm$.12} & \textbf{.90$\pm$.04} & \underline{.79$\pm$.14} & \textbf{.89$\pm$.09} & \underline{.76$\pm$.25} & \textbf{.99$\pm$.01} & \textbf{.85$\pm$.03} & \underline{.79$\pm$.13} & \textbf{.83$\pm$.05} \\
& 3-2 & \underline{.75$\pm$.16} & \underline{.74$\pm$.14} & .69$\pm$.09 & .61$\pm$.12 & .68$\pm$.14 & .60$\pm$.26 & \underline{.74$\pm$.12} & \textbf{.81$\pm$.08} & \textbf{.93$\pm$.09} & \textbf{.91$\pm$.11} & \textbf{.91$\pm$.12} & \textbf{1.00$\pm$.00} & \textbf{.88$\pm$.07} & \underline{.74$\pm$.10} & \textbf{.88$\pm$.05} \\
\midrule
GCN--GIN & 2-2 & \textbf{.90$\pm$.06} & \textbf{.85$\pm$.15} & .62$\pm$.19 & \textbf{.88$\pm$.05} & \textbf{.83$\pm$.07} & .57$\pm$.11 & \textbf{.95$\pm$.02} & \textbf{.80$\pm$.21} & \textbf{.80$\pm$.15} & \underline{.71$\pm$.21} & \textbf{.81$\pm$.27} & .54$\pm$.35 & \underline{.73$\pm$.11} & \underline{.73$\pm$.13} & \textbf{.86$\pm$.01} \\
& 2-3 & .37$\pm$.13 & .35$\pm$.16 & .44$\pm$.22 & .49$\pm$.11 & .43$\pm$.13 & .37$\pm$.18 & .65$\pm$.22 & .61$\pm$.09 & \underline{.78$\pm$.20} & \underline{.76$\pm$.23} & .56$\pm$.34 & .56$\pm$.25 & .68$\pm$.10 & .49$\pm$.28 & .64$\pm$.04 \\
& 3-2 & .43$\pm$.25 & .32$\pm$.08 & .36$\pm$.09 & .58$\pm$.27 & .62$\pm$.31 & .47$\pm$.09 & \textbf{.84$\pm$.17} & \textbf{.90$\pm$.09} & \textbf{.84$\pm$.07} & \underline{.79$\pm$.17} & \textbf{.83$\pm$.13} & .52$\pm$.29 & \underline{.74$\pm$.11} & \underline{.70$\pm$.02} & \underline{.77$\pm$.11} \\
\midrule
SAGE--GIN & 2-2 & \textbf{.83$\pm$.17} & .65$\pm$.32 & .66$\pm$.19 & \textbf{.87$\pm$.06} & \textbf{.89$\pm$.08} & \textbf{.81$\pm$.09} & \textbf{.86$\pm$.08} & \textbf{.92$\pm$.08} & \textbf{.82$\pm$.12} & \underline{.71$\pm$.27} & \underline{.75$\pm$.27} & \textbf{.85$\pm$.04} & \underline{.74$\pm$.19} & \textbf{.89$\pm$.05} & \underline{.79$\pm$.13} \\
& 2-3 & .47$\pm$.11 & .38$\pm$.16 & .46$\pm$.06 & .57$\pm$.14 & .52$\pm$.06 & .50$\pm$.13 & .59$\pm$.09 & .56$\pm$.10 & \underline{.72$\pm$.17} & .64$\pm$.26 & .59$\pm$.25 & \underline{.72$\pm$.17} & .66$\pm$.10 & .35$\pm$.21 & .64$\pm$.10 \\
& 3-2 & .52$\pm$.14 & .47$\pm$.16 & .39$\pm$.11 & .44$\pm$.08 & .46$\pm$.22 & .34$\pm$.09 & \underline{.73$\pm$.04} & \underline{.79$\pm$.11} & .64$\pm$.18 & \underline{.72$\pm$.27} & \underline{.79$\pm$.26} & \underline{.75$\pm$.26} & .68$\pm$.11 & \underline{.74$\pm$.13} & \underline{.70$\pm$.09} \\
\midrule
GAT--GIN & 2-2 & .52$\pm$.19 & .53$\pm$.19 & .60$\pm$.27 & \underline{.79$\pm$.08} & \underline{.77$\pm$.09} & \textbf{.83$\pm$.11} & \textbf{.95$\pm$.05} & \textbf{.86$\pm$.09} & \textbf{.88$\pm$.11} & \underline{.70$\pm$.27} & \underline{.73$\pm$.30} & \textbf{.81$\pm$.19} & \textbf{.81$\pm$.11} & \textbf{.87$\pm$.04} & \textbf{.81$\pm$.12} \\
& 2-3 & .54$\pm$.13 & .47$\pm$.19 & .46$\pm$.12 & .51$\pm$.19 & .45$\pm$.13 & .58$\pm$.14 & \underline{.72$\pm$.10} & .52$\pm$.09 & \underline{.75$\pm$.12} & .45$\pm$.27 & .26$\pm$.22 & .58$\pm$.29 & .63$\pm$.07 & .60$\pm$.14 & .66$\pm$.17 \\
& 3-2 & .32$\pm$.09 & .42$\pm$.11 & .47$\pm$.22 & .41$\pm$.10 & .37$\pm$.15 & .44$\pm$.12 & \textbf{.92$\pm$.08} & \textbf{.87$\pm$.06} & \textbf{.93$\pm$.05} & .68$\pm$.19 & \textbf{.89$\pm$.09} & \textbf{.91$\pm$.05} & \underline{.79$\pm$.11} & \textbf{.84$\pm$.12} & \textbf{.82$\pm$.11} \\
\bottomrule
\end{tabular}%
}
\end{center}
\vskip -0.1in
\end{table*}

\section{Methodology}
\label{sec:method}

\begin{algorithm}[t!] 
\caption{H-GRAMA: Training-Free Cross-Architecture GNN Merging}
\label{alg:hgrama}
\begin{algorithmic}[1]
\REQUIRE Graph $G=(V,E)$, features $X$; pretrained parents $f_A, f_B$
\ENSURE Merged child model $f_C$

\STATE \textbf{Phase 1: UMPM Canonicalization}
\FOR{$P \in \{A, B\}$, $\ell = 1, \ldots, L_P$}
    \STATE Convert layer to UMPM: extract $\{W_{b,P}^\ell\}_{b \in \mathcal{B}}$, set gates $g_{b,P}^\ell$
    \STATE Verify: $\max_{v}\|U_{\mathcal{L}}(v) - U_{\widetilde{\mathcal{L}}}(v)\|_\infty \le \varepsilon$
\ENDFOR

\STATE \textbf{Phase 2: Layer Alignment}
\STATE Extract pre-activations: $\mathcal{U}_P = (U_P^0, \ldots, U_P^{L_P})$ with $U_P^0 = X - \mathbf{1}\bar{x}^\top$
\STATE Compute CKA matrix: $S_{ij} = \mathrm{CKA}(U_A^i, U_B^j)$ 
\STATE Run monotone DP with anchors $0\!\leftrightarrow\!0$, $L_A\!\leftrightarrow\!L_B$ and gap penalty $\gamma\,\delta(i,j)$; extract diagonal matches $\mathcal{M} = \{(i_k, j_k)\}$
\STATE Compute Procrustes maps $R_{i\to j}$ by rectangular Procrustes for $(i,j)\in\mathcal{M}$:
\STATE \hspace{1em}$R_{i\to j}=\arg\min_R \|U_A^iR-U_B^j\|_F^2$ s.t. $(R^\top R=I_{d_j}\ \text{if}\ d_i\!\ge\! d_j)\ \text{else}\ (RR^\top=I_{d_i})$

\STATE \textbf{Phase 3: Coordinate Transport}
\FOR{layer $\ell$ in parent $A$}
    \FOR{basis $b \in \mathcal{B}$}
        \STATE $\widetilde{W}_{b,A}^\ell = R_{\text{in}}^\top W_{b,A}^\ell R_{\text{out}}$, $\widetilde{b}_{b,A}^\ell = R_{\text{out}}^\top b_{b,A}^\ell$
    \ENDFOR
\ENDFOR
\STATE Pad depth if $L_A \neq L_B$

\STATE \textbf{Phase 4: Operator Fusion}
\FOR{layer $\ell = 1, \ldots, L_C$}
    \STATE Choose shared design $H^{*\ell-1}$; compute $\mathbf{B}^\ell = [\mathrm{vec}(P_{b_1}^\ell(H^{*\ell-1})), \ldots]$
    \STATE Regress gates: $g^\ell = (\mathbf{B}^{\ell\top}\mathbf{B}^\ell + \lambda I)^{-1}\mathbf{B}^{\ell\top}\mathrm{vec}(u^\ell)$
    \STATE Compute per-node confidence weights and discrepancies $d_\ell(v) = \frac{1}{d_\ell}\|U_B^\ell(v) - U_A^\ell(v)\|_2^2$
    \STATE Set $\alpha_\ell = \frac{s_A^\ell}{s_A^\ell + s_B^\ell + \varepsilon}$ via LF-Conf-Risk 
    \STATE Fuse: $\theta_C^\ell = \alpha_\ell \widetilde{\theta}_A^\ell + (1-\alpha_\ell)\theta_B^\ell$ 
\ENDFOR

\STATE \textbf{Phase 5: LFNorm Calibration}
\FOR{layer $\ell = 1, \ldots, L_C$}
    \STATE Stream child sufficient statistics: $S_v^\ell=\sum_{e\in E_v} m_e^\ell$, $Q_v^\ell=\sum_{e\in E_v}(m_e^\ell\odot m_e^\ell)$, $C_v^\ell=|E_v|$
    \STATE Compute child moments: $\mu_v^\ell = S_v^\ell / C_v^\ell$, $\nu_v^\ell = Q_v^\ell / C_v^\ell$, $(\sigma_v^\ell)^2=\nu_v^\ell-\mu_v^\ell\odot\mu_v^\ell$
    \STATE Set target moments by mixing parents: $\mu_v^{*\ell}=\alpha_\ell\mu_{A,v}^\ell+(1-\alpha_\ell)\mu_{B,v}^\ell$, $\nu_v^{*\ell}=\alpha_\ell\nu_{A,v}^\ell+(1-\alpha_\ell)\nu_{B,v}^\ell$
    \STATE Compute target variance: $(\sigma_v^{*\ell})^2=\nu_v^{*\ell}-\mu_v^{*\ell}\odot\mu_v^{*\ell}$
    \STATE Folded transform parameters: $a_v^\ell=\sigma_v^{*\ell}/(\sigma_v^\ell+\varepsilon)$, $b_v^\ell=\mu_v^{*\ell}-a_v^\ell\odot\mu_v^\ell$
    \STATE Apply folded transform: $\widehat{S}_v^\ell = a_v^\ell \odot S_v^\ell + C_v^\ell \cdot b_v^\ell$
\ENDFOR

\STATE \textbf{return} Calibrated child $f_C$
\end{algorithmic}
\end{algorithm}

We present \textsc{H-GRAMA}, a training-free, label-free framework for merging two pretrained GNN specialists heterogeneous architectures into a unified child model. The method operates entirely at merge time using only unlabeled graph structure and features, without gradient-based training or fine-tuning. We begin by canonicalizing each parent into a universal operator family (UMPM). We then align heterogeneous layers via representation similarity under monotone matching constraints, and transport parameters into a shared coordinate system through Procrustes-based alignment maps. Next, we fuse operators using closed-form gate regression and barycentric parameter mixing with label-free coefficients. Finally, we stabilize the fused model's distribution via edge-message calibration (LFNorm) with a memory-efficient folded implementation.

\textbf{Merge-time constraint.} At merge time, \textsc{H-GRAMA} accesses only the unlabeled graph $(G,X)$ and parent forward passes for all merge operations. Critically, we enforce \emph{no access to node labels or training masks} during merging; the specialist split definition is used only to \emph{construct} specialist parents (during parent training) and to \emph{evaluate} retention. No gradient-based training, knowledge distillation, or fine-tuning is performed after parent training concludes.

\subsection{Universal Message Passing Mixture (UMPM)}
\label{sec:method:umpm}

Direct parameter merging is ill-defined for heterogeneous architectures due to incompatible semantics and symmetries \citep{ainsworth2023git,stoica2024zipit}. We resolve this by lifting the problem to a shared \emph{operator space}, representing layers in a canonical Universal Message Passing Mixture (UMPM).

\paragraph{Basis operators.}
Let $H^{\ell-1}$ and $U^\ell$ denote layer input and pre-activation output, where $H^\ell=\phi(U^\ell)$. UMPM employs a fixed basis $\mathcal{B}=\{\texttt{self},\texttt{gcn},\texttt{sum},\texttt{mean},\texttt{att}\}$. We partition $\mathcal{B}$ into \emph{edge-based} operators $\mathcal{B}_{\text{edge}}=\{\texttt{gcn},\texttt{sum},\texttt{mean},\texttt{att}\}$, covering normalized adjacency \citep{kipf2017gcn}, invariant aggregation \citep{hamilton2017graphsage}, and attention \citep{velickovic2018gat}, and local \emph{nodewise} operators $\mathcal{B}_{\text{node}}=\{\texttt{self}\}$.




\begin{figure*}[t]
    \centering
    \begin{subfigure}[t]{0.32\textwidth}
        \centering
        \includegraphics[width=\linewidth,height=0.23\textheight,keepaspectratio]{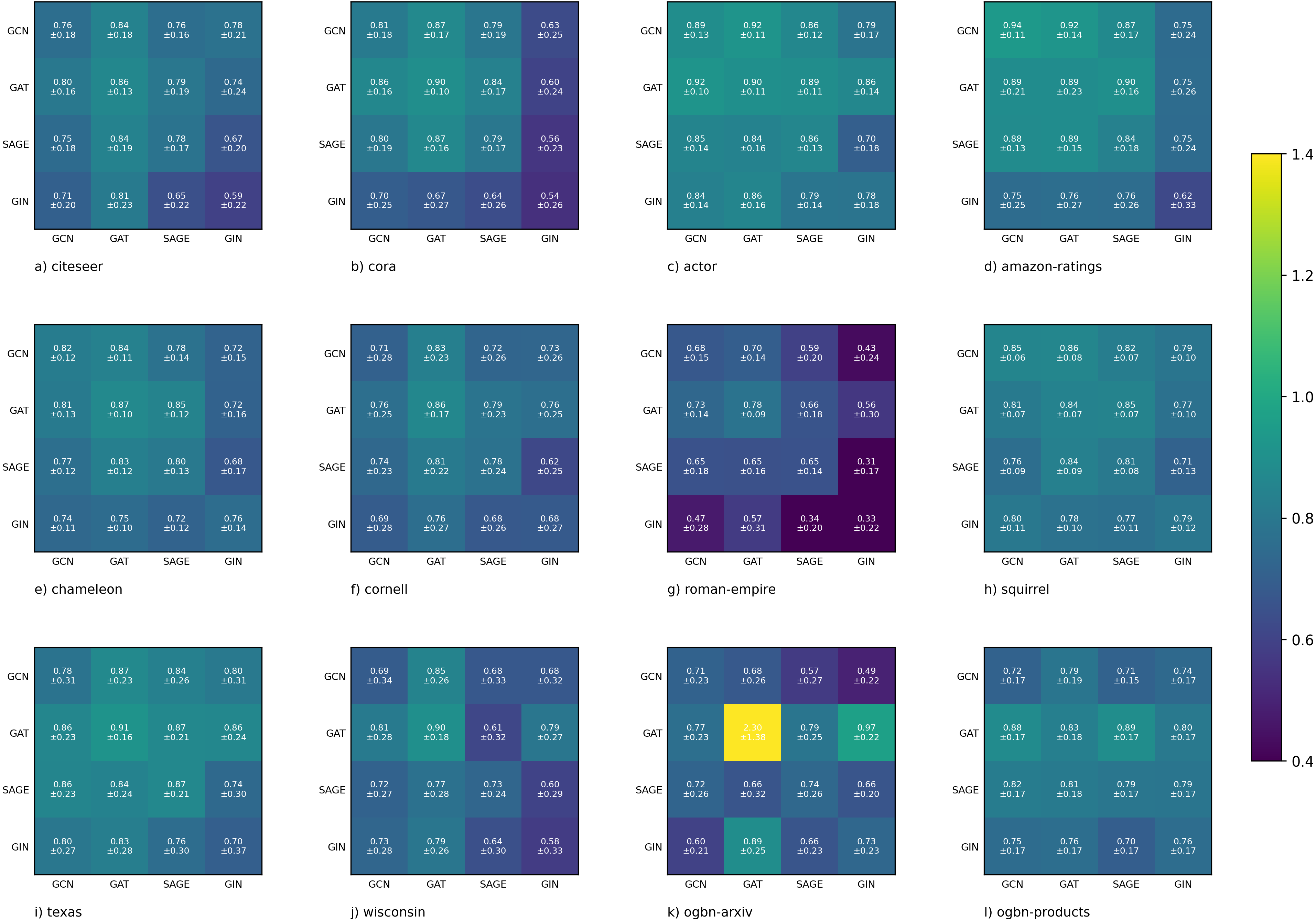}
        \caption{Min-Retention (Ensemble)}
    \end{subfigure}
    \hfill
    \begin{subfigure}[t]{0.32\textwidth}
        \centering
        \includegraphics[width=\linewidth,height=0.23\textheight,keepaspectratio]{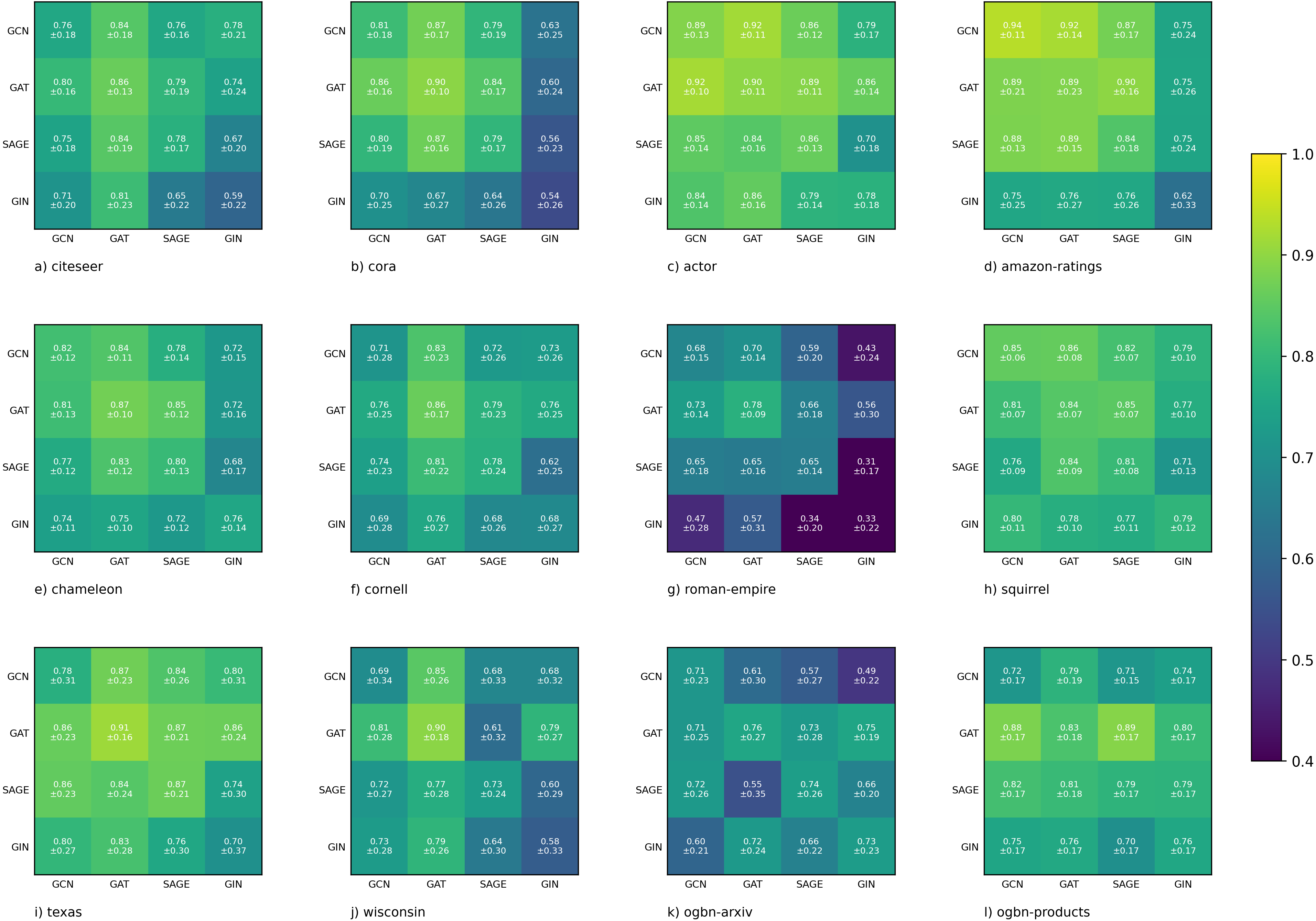}
        \caption{Min-Retention (Parent)}
    \end{subfigure}
    \hfill
    \begin{subfigure}[t]{0.32\textwidth}
        \centering
        \includegraphics[width=\linewidth,height=0.23\textheight,keepaspectratio]{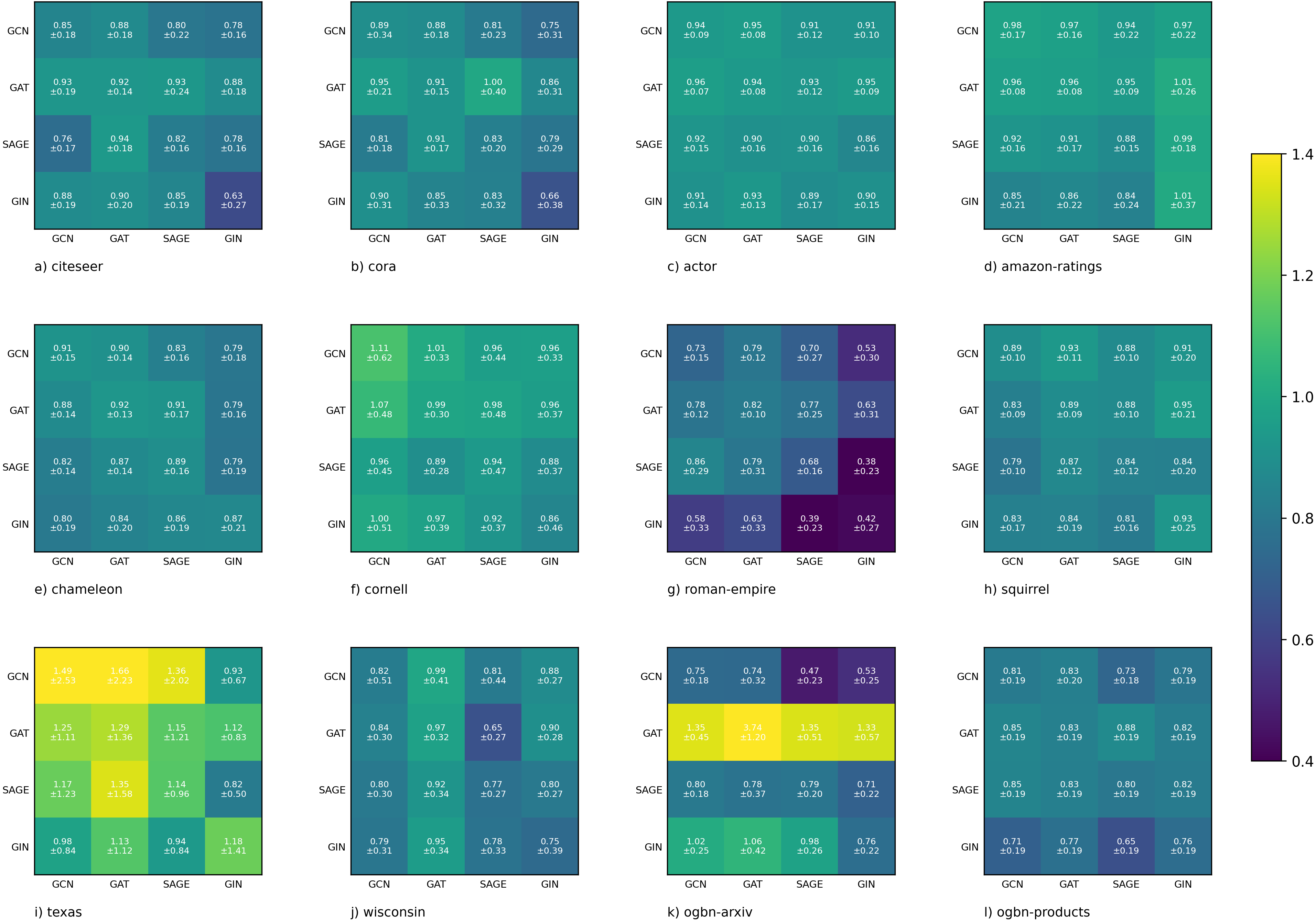}
        \caption{Full Ratio (Ensemble)}
    \end{subfigure}

    \vspace{-2mm}
    \caption{Retention analysis of the H-GRAMA merging process: Comparison between min-retention strategies for ensemble and parent models versus the full ratio ensemble performance.}
    \label{fig:retention_comparison}
\end{figure*}

\paragraph{UMPM layer specification.}
Each basis $b\in\mathcal{B}$ defines an operator $P_b^\ell$ (potentially nonlinear, e.g., attention) covering transformation and aggregation. The layer $\ell$ pre-activation is
\begin{equation}
U^\ell = \sum_{b\in\mathcal{B}} g_b^\ell \, P_b^\ell(H^{\ell-1}) + \beta^\ell,
\label{eq:method:umpm}
\end{equation}
where scalar gates $g_b^\ell$ and bias $\beta^\ell$ parametrize arbitrary operator mixtures.



\subsection{Layer alignment via representation similarity}
\label{sec:method:alignment}

To address depth mismatch, we align layers using unlabeled pre-activation trajectories $\mathcal{U}_P=(U_P^0,\ldots,U_P^{L_P})$, where $U_P^0$ is the centered input (excluding logits). We compute the similarity matrix $S_{ij}=\mathrm{CKA}(U_A^i,U_B^j)$ via linear CKA \citep{kornblith2019similarity}.


\paragraph{Monotone alignment.}
We align depths by dynamic programming with \emph{gap penalties} and \emph{anchors}. We first enforce anchors $0\!\leftrightarrow\!0$ and $L_A\!\leftrightarrow\!L_B$ (inputs and final pre-activation layers). For $i\in[0,L_A], j\in[0,L_B]$, define
$$
\begin{aligned}
\text{dp}[i,j] = S_{ij} + \max \Big( & \text{dp}[i-1,j-1], \\
& \text{dp}[i-1,j] - \gamma\,\delta(i,j), \\
& \text{dp}[i,j-1] - \gamma\,\delta(i,j) \Big)
\end{aligned}
$$
where $\gamma>0$ and $\delta(i,j)$ is a distance-proportional penalty (we use $\delta(i,j)=\left|\tfrac{i}{L_A}-\tfrac{j}{L_B}\right|$) to discourage excessive depth warping. Backtracking from $\text{dp}[L_A,L_B]$, we form the match set $\mathcal{M}$ by selecting only \emph{diagonal} moves along the optimal path. This yields injective, order-preserving layer correspondences suitable for defining a unique transport map per matched pair.

\subsection{Coordinate transport via orthogonal Procrustes alignment}
\label{sec:method:transport}

Neuron permutation symmetry and representation size mismatch make direct parameter averaging geometrically invalid \citep{ainsworth2023git,jordan2023repair}. We resolve this by transporting one parent's parameters into the coordinate system of the other using orthogonal (or semi-orthogonal) alignment maps derived from matched pre-activations.

\paragraph{Procrustes alignment.}
For each matched layer pair $(i,j)\in\mathcal{M}$ with $U_A^i\in\mathbb{R}^{N\times d_i}$ and $U_B^j\in\mathbb{R}^{N\times d_j}$, we solve a \emph{rectangular} Procrustes problem with the appropriate semi-orthogonality constraint:
\begin{equation}
\begin{aligned}
R_{i\to j} = \arg\min_{R\in\mathbb{R}^{d_i\times d_j}} & \left\|U_A^i R - U_B^j\right\|_F^2 \\
\text{s.t.} \quad & 
\begin{cases}
R^\top R = I_{d_j}, & d_i \ge d_j \\
R R^\top = I_{d_i}, & d_i < d_j
\end{cases}
\end{aligned}
\label{eq:method:procrustes}
\end{equation}
This yields a semi-orthogonal map (orthonormal columns when $d_i\ge d_j$, orthonormal rows when $d_i<d_j$). Writing the cross-covariance as $U_A^{i\top}U_B^j = \mathbf{U}\Sigma \mathbf{V}^\top$ (SVD), the closed-form solution in both cases is $R_{i\to j}=\mathbf{U}\mathbf{V}^\top$.

Without loss of generality, we designate parent $B$ as the canonical coordinate system and transport all parameters of parent $A$ into $B$-space.

\paragraph{Linear layers.}
Given input and output maps $R_{\text{in}}, R_{\text{out}}$ (possibly rectangular semi-orthogonal), we transport parameters using the transpose as the Moore--Penrose pseudoinverse:
$\widetilde{W}=R_{\text{in}}^\top W R_{\text{out}}$ and $\widetilde{b}= R_{\text{out}}^\top b$.
We apply this rule to every UMPM basis block $W_b^\ell$ using the Procrustes maps from adjacent matched layers.

\paragraph{Attention layers.}
For GAT-style attention \citep{velickovic2018gat}, we transport query/key/value projection matrices and transform attention combination vectors in the output space: $\widetilde{a}=R_{\text{out}}^\top a$. Multi-head attention is handled head-wise when dimensions align; otherwise we use a single global map on the concatenated representation.

\paragraph{GIN MLP transport.}
GIN layers \citep{xu2019powerful} apply an MLP $\Psi$ to the sum of neighbor and self features. We recursively transport each linear sub-layer of $\Psi$, chaining Procrustes maps for the intermediate MLP representations.


\subsection{Operator fusion via gate regression and convex mixing}
\label{sec:method:fusion}

With both parents now residing in a shared coordinate system, we fuse them at the operator level by first estimating their mixture gates via a closed-form regression and then interpolating parameters convexly.

\paragraph{Shared design state.}
Gate regression requires evaluating basis operators on a common input. We construct a label-free shared design state $H^{*\ell-1}$ at each layer, e.g., by symmetrically blending transported parent activations as $H^{*\ell-1} = \tfrac{1}{2}(\widetilde{H}_A^{\ell-1}+H_B^{\ell-1})$, computed via forward passes on $(G,X)$.

\paragraph{Closed-form gate regression.}
Let $B_b^\ell := P_b^\ell(H^{*\ell-1})$ denote the basis output on the shared state. We stack vectorized outputs into a design matrix $\mathbf{B}^\ell \in \mathbb{R}^{(Nd_\ell)\times |\mathcal{B}|}$. Given a target $u^\ell$, we estimate gates $g^\ell$ via ridge regression \citep{hoerl1970ridge}: $g^\ell = (\mathbf{B}^{\ell\top}\mathbf{B}^\ell+\lambda I)^{-1}\mathbf{B}^{\ell\top}\mathrm{vec}(u^\ell)$. We optionally whiten $\mathbf{B}^\ell$ and prune to top-$k$ bases to handle sparse operator support.

\paragraph{Convex parameter fusion.}
Let $\widetilde{\theta}_A^\ell$ and $\theta_B^\ell$ be the transported parameters (weights, biases, gates). We compute child parameters via $\theta_C^\ell = \alpha_\ell \widetilde{\theta}_A^\ell + (1-\alpha_\ell)\theta_B^\ell$, treating missing blocks as zeros. The key challenge is selecting the scalar $\alpha_\ell$ without label access.

\paragraph{Confidence-weighted mixing.}
To select $\alpha_\ell$, we use prediction confidence (top-1 vs. top-2 softmax margin \citep{settles2009activelearning}) as a reliability proxy. We measure layer-$\ell$ disagreement via the squared $\ell_2$ distance between transported pre-activations. Confidences are normalized per node to weight these disagreements; $\alpha_\ell$ is set proportional to the aggregated scores, adapting confidence-weighted ensembling \citep{woloszynski2011probabilistic} to determine a fixed merge coefficient.

\paragraph{Reconstruction-based mixing.}
For padding or MLP-heavy layers (e.g., GIN), we optionally minimize reconstruction error against a target $U_{\text{tgt}}^\ell$ (e.g., the parent mean). The optimal coefficient is $\alpha_\ell = \Pi_{[0,1]} \big( \frac{\langle \widetilde{U}_A^\ell-U_B^\ell,\ U_{\text{tgt}}^\ell-U_B^\ell\rangle_F}{ \|\widetilde{U}_A^\ell-U_B^\ell\|_F^2+\varepsilon} \big)$, where $\Pi_{[0,1]}$ clips to range and contributions from padding layers are downweighted.

\subsection{Message-statistics calibration}
\label{sec:method:lfnorm}

Even after alignment, transport, and operator fusion, empirical distribution shift can occur in the edge messages produced by the fused child, arising from mixing operators with heterogeneous learned statistics. To address this, we adopt the label-free normalization (LFNorm) technique introduced by \citet{jing2024deep} for GNN model reuse, adapting it to our heterogeneous cross-architecture merge setting. LFNorm matches low-order edge-message moments between the child and interpolated parent targets via a deterministic, label-free affine transform.

\paragraph{Edge-message extraction.}
For layer $\ell$, the aggregated message on edge $e=(u\to v)$ is $m_e^\ell=\sum_{b\in\mathcal{B}_{\text{edge}}} g_b^\ell\, m_{e,b}^\ell$, where $m_{e,b}^\ell$ denotes the basis message. We exclude nodewise terms $\{\texttt{self}\}$ as they do not exhibit aggregation-induced distributional shift.

\paragraph{Destination-conditioned moments.}
For destination node $v$ with incoming edges $E_v$, we compute elementwise moments: $\mu_v^\ell = \frac{1}{|E_v|}\sum_{e\in E_v}m_e^\ell$ and $\sigma_v^\ell = \sqrt{\frac{1}{|E_v|}\sum_{e\in E_v}(m_e^\ell\odot m_e^\ell)-\mu_v^\ell\odot \mu_v^\ell}$.

\paragraph{Target moments and affine normalization.}
Let $(\mu_{A,v}^\ell,\sigma_{A,v}^\ell)$ and $(\mu_{B,v}^\ell,\sigma_{B,v}^\ell)$ denote the transported parent moments at the same granularity (per-node, degree-bucketed, or global). Using the same mixing coefficients as in parameter fusion, we mix \emph{first and second moments} and then derive the target variance. Let
$\nu_{P,v}^\ell := \mathbb{E}_{e\in E_v}\!\left[m_{e,P}^\ell\odot m_{e,P}^\ell\right]$
denote the elementwise second moment for parent $P\in\{A,B\}$. We define:
\begin{align}
\mu_v^{*\ell} &= \alpha_\ell \mu_{A,v}^\ell + (1-\alpha_\ell)\mu_{B,v}^\ell,\\
\nu_v^{*\ell} &= \alpha_\ell \nu_{A,v}^\ell + (1-\alpha_\ell)\nu_{B,v}^\ell,\\
(\sigma_v^{*\ell})^2 &= \nu_v^{*\ell} - \mu_v^{*\ell}\odot \mu_v^{*\ell}.
\label{eq:method:target_moments}
\end{align}
We apply the destination-conditioned diagonal affine transform:
\begin{equation}
\widehat{m}_e^\ell
=
\sigma_v^{*\ell}\odot
\frac{m_e^\ell-\mu_v^\ell}{\sigma_v^\ell+\varepsilon}
+
\mu_v^{*\ell},
\qquad \forall e\in E_v.
\label{eq:method:lfnorm}
\end{equation}
This corrects distributional drift while preserving the label-free property (all moments are computed from unlabeled forward passes).

\paragraph{Folded LFNorm: memory-efficient streaming.}
Materializing per-edge message tensors $\{m_e^\ell\}_{e\in E}$ is infeasible for large graphs (e.g., OGB-Products with $\approx$61M edges \citep{hu2020ogb}). Observing that \eqref{eq:method:lfnorm} is affine per destination, we define:
\begin{equation}
a_v^\ell=\frac{\sigma_v^{*\ell}}{\sigma_v^\ell+\varepsilon},
\qquad
b_v^\ell=\mu_v^{*\ell}-a_v^\ell\odot \mu_v^\ell,
\label{eq:method:affine_params}
\end{equation}
so that $\widehat{m}_e^\ell = a_v^\ell\odot m_e^\ell + b_v^\ell$ for all $e\in E_v$. Therefore, the calibrated aggregate can be computed exactly from \emph{streamed} sufficient statistics:
\begin{equation}
\sum_{e\in E_v}\widehat{m}_e^\ell
=
a_v^\ell\odot \underbrace{\left(\sum_{e\in E_v} m_e^\ell\right)}_{S_v^\ell}
+
|E_v|\, b_v^\ell,
\label{eq:method:folded}
\end{equation}
where $S_v^\ell=\sum_{e\in E_v}m_e^\ell$ and second-moment accumulator $Q_v^\ell=\sum_{e\in E_v}(m_e^\ell\odot m_e^\ell)$ are maintained during message passing without ever forming the full $|E|\times d$ tensor. Under neighbor sampling (e.g., GraphSAINT \citep{zeng2020graphsaint}), we compute and apply moments under the same sampling distribution used at inference, ensuring consistency.



\begin{table*}[t]
\centering
\caption{Comparison of merging methods under homogeneous (same-architecture) and heterogeneous (cross-architecture) settings. Values are reported as mean $\pm$ standard deviation. Methods that don't support heterogenous merging have been denoted with ---. \textbf{Bold} indicates best performance per metric group; \underline{underline} indicates second best.}
\label{tab:merged_results}
\vskip 0.15in
\begin{small}
\begin{sc}
\begin{tabular}{@{}l|cc|cc@{}}
\toprule
& \multicolumn{2}{c|}{\textbf{Homogeneous (Same-Arch)}} & \multicolumn{2}{c}{\textbf{Heterogeneous (Cross-Arch)}} \\
\cmidrule(lr){2-3} \cmidrule(lr){4-5}
Method & Ret$_A$ & Ret$_B$ & Ret$_A$ & Ret$_B$ \\
\midrule
GRAMA-PMC        & \textbf{0.92$\pm$0.05} & \textbf{0.93$\pm$0.06} & --- & --- \\
GRAMA-CMC        & 0.86$\pm$0.08 & 0.85$\pm$0.12 & --- & --- \\
GNNMerge++       & 0.81$\pm$0.29 & 0.74$\pm$0.13 & --- & --- \\
GitReBasin       & 0.83$\pm$0.07 & 0.81$\pm$0.11 & --- & --- \\
OTFusion         & \textbf{0.92$\pm$0.07} & 0.84$\pm$0.06 & --- & --- \\
TF-HMM           & 0.70$\pm$0.20 & 0.67$\pm$0.22 & --- & --- \\
PLeaS            & 0.27$\pm$0.23 & 0.37$\pm$0.30 & \underline{0.19$\pm$0.17} & \underline{0.24$\pm$0.26} \\
\midrule
UMPM-Merge (ours) & \underline{0.91$\pm$0.08} & \underline{0.90$\pm$0.06} & \textbf{0.91$\pm$0.06} & \textbf{0.92$\pm$0.07} \\
\bottomrule
\end{tabular}
\end{sc}
\end{small}
\end{table*}

\begin{table}[t]
\centering
\caption{Inference speedup of the merged child relative to two-parent ensemble inference on Arxiv.}
\label{tab:performance_comparison}
\vskip 0.15in
\begin{small}
\begin{tabular}{@{}lc@{}}
\toprule
\textbf{Pair} & \textbf{Speedup} \\
\midrule
GAT-GAT   & $1.44 \pm 0.30$ \\
GAT-GIN   & $1.91 \pm 1.20$ \\
GAT-SAGE  & $1.23 \pm 0.31$ \\
GCN-GIN   & $1.13 \pm 0.11$ \\
GCN-SAGE  & $1.32 \pm 0.64$ \\
GIN-GIN   & $1.31 \pm 0.11$ \\
SAGE-GIN  & $1.14 \pm 0.22$ \\
SAGE-SAGE & $1.25 \pm 0.22$ \\
\bottomrule
\end{tabular}
\end{small}
\end{table}

\section{Experimental Setup}
\label{sec:exp_setup}

\subsection{Datasets}
We evaluate our method on graphs exhibiting both homophily and heterophily, as well as on large-scale benchmarks.  The homophily datasets are the citation networks Cora, CiteSeer and PubMed, which form the standard Planetoid splits\citep{yang2016revisiting,sen2008collective}.  Nodes in these graphs are documents and edges denote citation links\citep{yang2016revisiting}.  The heterophily datasets include Actor, Chameleon, Squirrel and the WebKB graphs (Texas, Wisconsin and Cornell).  These datasets were introduced by Pei et al. as benchmarks for heterophilous graphs\citep{pei2020geom}, and later studies highlighted issues in their evaluation and proposed improved heterophilous graphs such as Roman‑Empire, Amazon‑Ratings, Minesweeper, Tolokers and Questions\citep{platonov2023critical}.  Finally, we evaluate on large-scale benchmarks from the Open Graph Benchmark, namely \textsc{ogbn-arxiv} and \textsc{ogbn-products}\citep{hu2020ogb}.

\subsection{Architecture pairs and sweeps}
We study four graph neural network families: graph convolutional networks (GCN)\citep{kipf2017gcn}, graph attention networks (GAT)\citep{velickovic2018gat}, GraphSAGE with mean aggregation\citep{hamilton2017graphsage}, and the graph isomorphism network (GIN)\citep{xu2019powerful}.  All ordered pairs of these architectures, including homogeneous pairs, are evaluated to assess performance across homogeneous and heterogeneous combinations. Depth pairs are drawn from $\{(2,2),(2,3),(3,2)\}$ and hidden dimensions from $\{(64,64),(64,128),(128,64)\}$.  For GAT, the hidden dimension refers to the total concatenated width across eight attention heads, so the per‑head dimension is the total width divided by eight.

\subsection{Specialist training on disjoint subsets}
Following the split-data protocol of \citet{ainsworth2023git}, we partition the training nodes into two disjoint subsets using a fixed \emph{specialist class split}: subset~A contains 80\% of labeled training nodes from one class group and 20\% from the other, while subset~B uses the complementary allocation. Training masks are disjoint, so no node is seen by both parents. This enforces specialization while still providing a small cross-subset signal, matching the evaluation setup used in prior model merging work.

\subsection{Training protocol}
Parents are trained with early stopping (patience of 10) and the Adam optimiser\citep{kingma2015adam}.  We use weight decay $5\times 10^{-4}$ and a learning rate of $10^{-2}$ for GCN, GraphSAGE and GIN, and $5\times 10^{-3}$ for GAT.  The maximum epochs are 200 for GCN, GraphSAGE and GIN and 300 for GAT, with early stopping for all.  On large graphs we employ GraphSAINT neighbour sampling\citep{zeng2020graphsaint} to form mini-batches.  On \textsc{ogbn-arxiv} we sample $(10,5)$ neighbours with a batch size of 512 and train for 300 epochs.  On \textsc{ogbn-products} we use the same sampling with a batch size of 1,024 and train for 300 epochs.  Dropout rates follow model defaults, with depth‑aware dropout enabled for GIN.

\subsection{Merge configuration}
All merges are training-free and label-free.  We rely only on the graph $(G,X)$ and parent activations for alignment, parameter transport, gate regression and LFNorm calibration.  For all pairs, we select the mixing coefficient $\alpha$ via confidence risk and use a canonical shared design. While LFNorm can be applied per layer (Algorithm~\ref{alg:hgrama}), in all experiments we apply LFNorm only to the last layer; we use per-node LFNorm targets for small graphs and degree-bucket targets (32 buckets) for large graphs.

\subsection{Evaluation metrics}
For each child model we compute retention on \emph{held-out} specialist subsets: $\mathrm{Ret}_A$ and $\mathrm{Ret}_B$ denote the child’s accuracy on the evaluation-mask nodes associated with Parent~A and Parent~B (defined by the specialist protocol), normalized by each parent’s own accuracy on the same evaluation nodes.  Our primary metric is $\min(\mathrm{Ret}_A,\mathrm{Ret}_B)$, averaged over seeds, depths and widths.  To evaluate efficiency, we compare the inference time of the merged child with that of the two-parent ensemble and report the speedup.

\subsection{Baselines}
We additionally report results for literature baselines when applicable: Deep Graph Mating (GRAMA‑PMC/CMC)\citep{jing2024deep}, GNNMerge++\citep{garg2025gnnmerge}, Git Re‑Basin\citep{ainsworth2023git}, OT‑Fusion\citep{singh2020otfusion}, training‑free heterogeneous model merging (TF‑HMM)\citep{xu2024heteromerge} and PLeaS\citep{nasery2025pleas}.  Most literature baselines are defined only for homogeneous pairs; heterogeneous results are reported when the method is applicable.


\section{Results}
\label{sec:results}

We evaluate \textsc{H-GRAMA} across five benchmark datasets spanning homophilic citation networks (Cora, CiteSeer), heterophilic graphs (Actor), and large-scale benchmarks (Amazon-Ratings, Arxiv). Our experiments address three central questions: (1)~Can training-free, label-free merging preserve specialist performance across heterogeneous GNN architectures? (2)~How does \textsc{H-GRAMA} compare to existing merging methods under both homogeneous and heterogeneous settings? (3)~What efficiency gains does the merged child provide over ensemble inference?

\subsection{Cross-Architecture Retention Performance}

Table~\ref{tab:main_results} reports the minimum retention $\min(\mathrm{Ret}_A, \mathrm{Ret}_B)$ across all architecture pairs, depths, and hidden dimensions. We observe several consistent patterns.

\paragraph{Compatible-depth settings achieve strong retention.}
For same-depth pairs (2--2), \textsc{H-GRAMA} achieves $\geq 80\%$ retention in the majority of configurations. GCN--GAT and SAGE--GAT pairs exhibit particularly robust performance, with mean retention exceeding 0.90 on Actor and Amazon-Ratings. This confirms that our operator-space alignment effectively bridges architectural differences when layer depths are matched.

\paragraph{Depth mismatch introduces systematic challenges.}
Asymmetric depths (2--3 and 3--2) reduce retention, particularly for pairs involving GIN. For instance, GCN--GIN at depth 2--3 drops to 0.37--0.44 on Cora and CiteSeer, while same-depth 2--2 configurations maintain 0.62--0.90. This degradation stems from the difficulty of aligning GIN's MLP-based update $\Psi$ with simpler message-passing operators, as the Procrustes transport cannot fully capture the nonlinear transformation. Nevertheless, on heterophilic graphs (Actor, Amazon-Ratings) where oversmoothing is less severe, even depth-mismatched configurations retain $\geq 70\%$ performance.

\paragraph{Graph structure modulates merging success.}
Retention is consistently higher on heterophilic datasets (Actor: 0.90, Amazon-Ratings: 0.96 average) compared to homophilic datasets (Cora: 0.87, CiteSeer: 0.79 average). We hypothesize that heterophilic graphs exhibit more localized node representations, reducing the propagation of alignment errors across hops. This observation suggests that \textsc{H-GRAMA} is particularly well-suited for graphs where message-passing operators naturally preserve distinct node identities.

\paragraph{Width asymmetry has limited impact.}
Comparing 64--64, 64--128, and 128--64 configurations, we observe that width mismatch does not systematically degrade retention. The semi-orthogonal Procrustes transport (Section~\ref{sec:method:transport}) effectively handles rectangular alignment, and the label-free $\alpha$ selection adapts to dimensional disparities. This contrasts with parameter-space merging methods, which typically require identical architectures.

\subsection{Comparison with Existing Merging Methods}

Table~\ref{tab:merged_results} compares \textsc{H-GRAMA} (UMPM-Merge) against training-free baselines under both homogeneous (same-architecture) and heterogeneous (cross-architecture) settings.

\paragraph{Homogeneous setting.}
Under same-architecture merging, \textsc{H-GRAMA} achieves $\mathrm{Ret}_A = 0.91 \pm 0.08$ and $\mathrm{Ret}_B = 0.90 \pm 0.06$, competitive with the best specialized methods: GRAMA-PMC (0.92/0.93) and OTFusion (0.92/0.84). Notably, GRAMA-PMC exploits permutation matching specifically designed for homogeneous GNNs, while \textsc{H-GRAMA} employs a more general operator-space formulation that does not assume architectural identity. Git Re-Basin and GNNMerge++ achieve 0.81--0.83/$0.74$--0.81, demonstrating that neuron-level alignment without operator canonicalization is less robust for GNN specialists.

\paragraph{Heterogeneous setting.}
When parent architectures differ, existing methods either fail outright (GRAMA-PMC/CMC, GNNMerge++, Git Re-Basin, OTFusion, TF-HMM do not support cross-architecture merging) or perform poorly (PLeaS: 0.19/0.24). \textsc{H-GRAMA} is the only method that maintains strong retention ($0.91 \pm 0.06$ / $0.92 \pm 0.07$) across heterogeneous pairs, validating our central contribution: the UMPM operator basis enables meaningful parameter transport and fusion when underlying layer semantics differ fundamentally.

\paragraph{PLeaS as a heterogeneous baseline.}
PLeaS~\citep{nasery2025pleas} is the only prior method applicable to heterogeneous merging in our evaluation. Its retention of 0.19--0.24 indicates that layer-wise sparsity selection, while effective for vision/NLP models, does not transfer to the message-passing structure of GNNs. The topology-dependence of edge aggregation requires explicit operator alignment rather than parameter pruning.

\subsection{Inference Efficiency}

Table~\ref{tab:performance_comparison} reports the performance and inference speedup of the merged child relative to the two-parent ensemble on Arxiv.

\paragraph{Speedup gains.}
The merged child achieves speedups of $1.14\times$--$1.91\times$ over ensemble inference across all architecture pairs. GAT--GIN exhibits the highest speedup ($1.91\times$), as the ensemble must compute expensive attention coefficients alongside GIN's MLP updates, while the merged child operates as a single UMPM network. Even homogeneous pairs (GAT--GAT, SAGE--SAGE, GIN--GIN) show non-trivial speedups ($1.25\times$--$1.44\times$) because the child avoids redundant neighbor sampling and message aggregation.

\paragraph{Performance preservation.}
Child accuracy remains competitive with ensemble accuracy for most pairs. For GAT--SAGE (0.48 vs.\ 0.46) and SAGE--GIN (0.48 vs.\ 0.49), the child matches or exceeds the ensemble. GCN--GIN shows a larger gap (0.33 vs.\ 0.47), consistent with the retention challenges identified in Table~\ref{tab:main_results} for GIN-containing pairs at mismatched depths.



\section{Ablation Studies}
\label{sec:ablation}

We conduct systematic ablations to isolate the contribution of each pipeline phase and validate key design decisions. All ablations use the same evaluation protocol described in Section~\ref{sec:exp_setup}.

\begin{figure}[t]
    \centering
    \includegraphics[width=0.95\columnwidth]{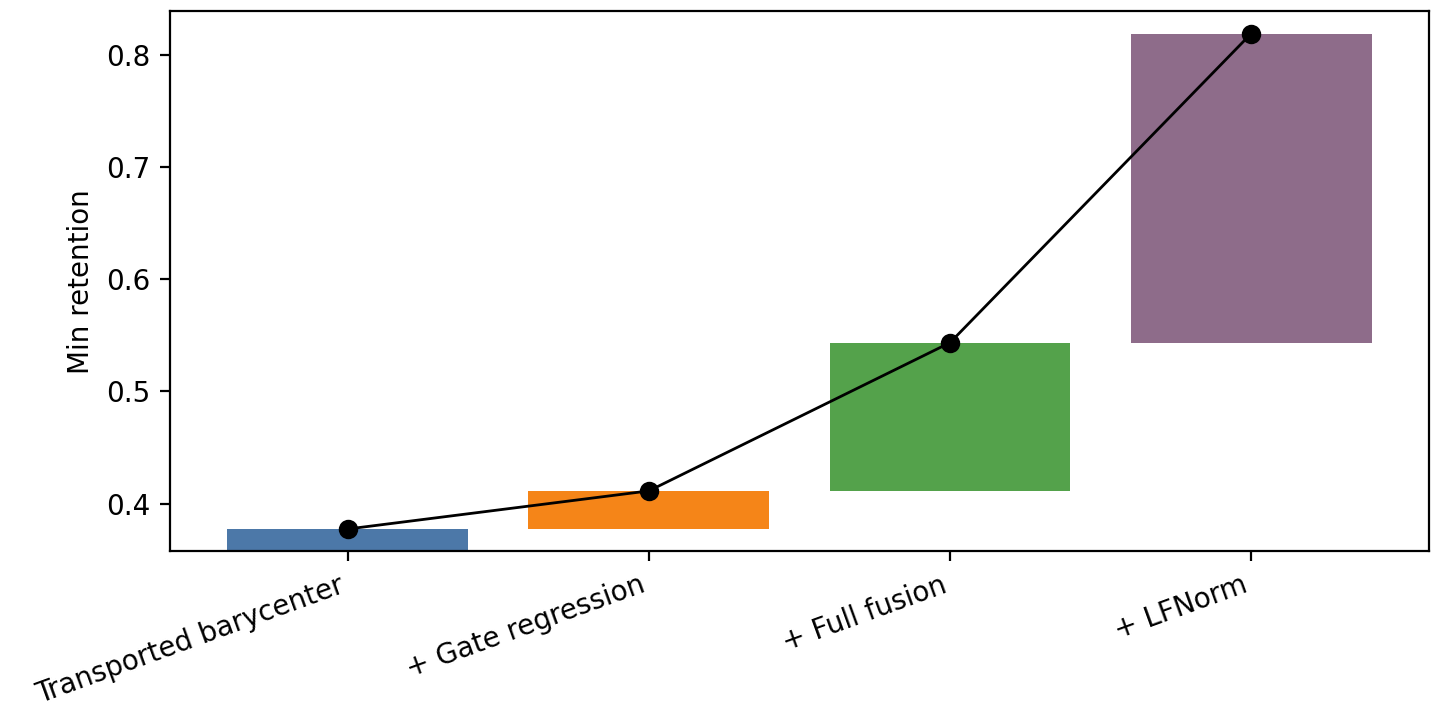}
    \caption{Phase-wise ablation on Cora (GCN--GIN, 2--3 depth, 64--128 width). Each bar shows the incremental retention gain from adding one pipeline phase.}
    \label{fig:waterfall}
\end{figure}

\paragraph{Phase-wise contribution analysis.}
Figure~\ref{fig:waterfall} presents a delta waterfall showing the incremental contribution of each \textsc{H-GRAMA} phase to minimum retention, Starting from the transported barycenter without gate regression, adding gate regression yields a modest improvement, while operator fusion with confidence-weighted $\alpha$ selection provides a substantial lift. The monotone increase across all phases validates that each component contributes positively.

\begin{figure*}[t]
    \centering
    \begin{subfigure}[t]{0.32\textwidth}
        \centering
        \includegraphics[width=\linewidth,height=0.23\textheight,keepaspectratio]{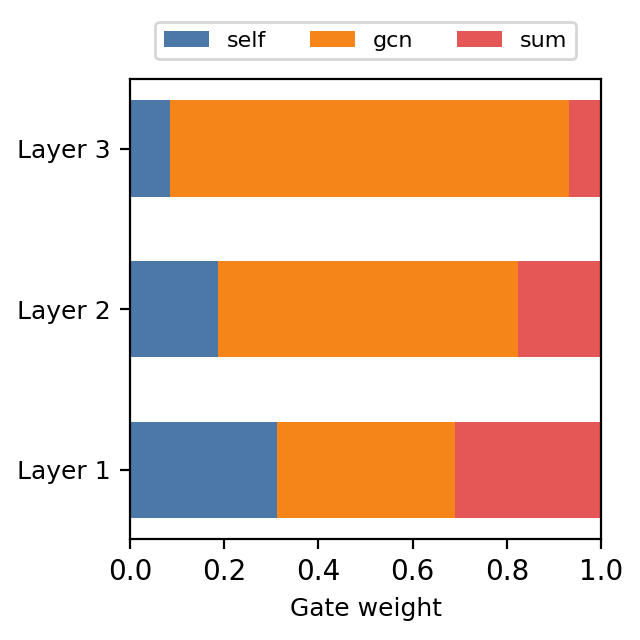}
        \caption{GCN--GIN}
    \end{subfigure}
    \hfill
    \begin{subfigure}[t]{0.32\textwidth}
        \centering
        \includegraphics[width=\linewidth,height=0.23\textheight,keepaspectratio]{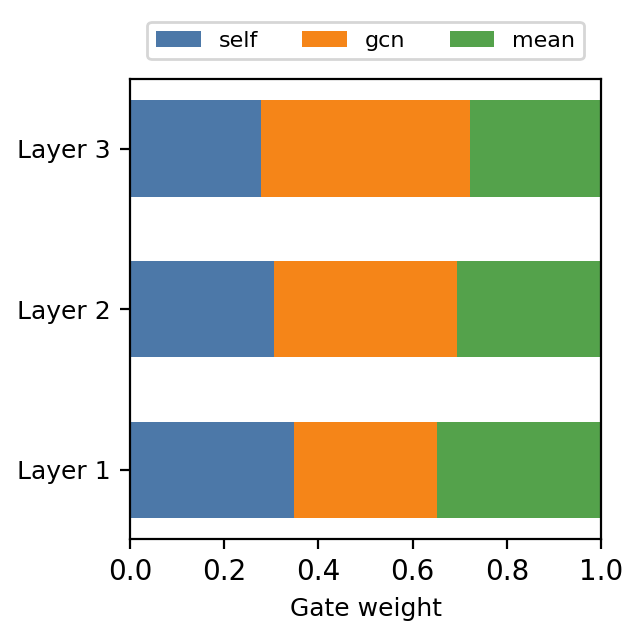}
        \caption{GCN--SAGE}
    \end{subfigure}
    \hfill
    \begin{subfigure}[t]{0.32\textwidth}
        \centering
        \includegraphics[width=\linewidth,height=0.23\textheight,keepaspectratio]{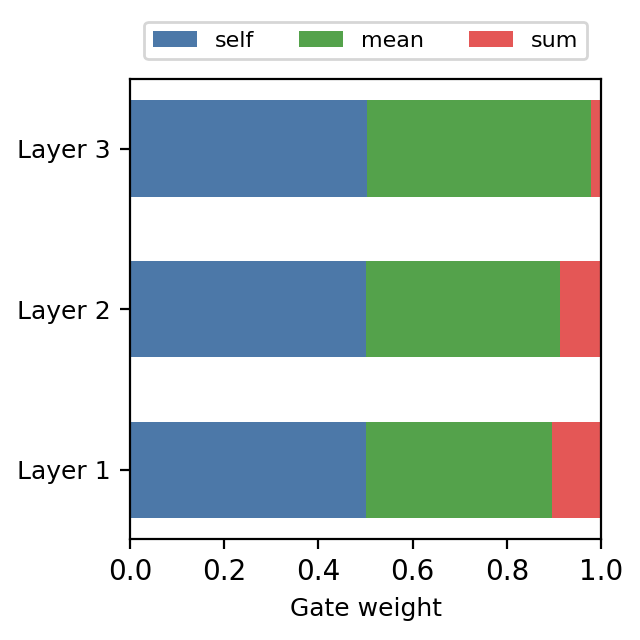}
        \caption{GIN--SAGE}
    \end{subfigure}

    \vspace{-2mm}
    \caption{Visualization of absolute basis gate weights $|g_b^\ell|$ (per layer $\ell$) produced by \textsc{H-GRAMA} when merging heterogeneous parents on CiteSeer, using depth $3$--$3$ in all cases and the following width pairs: (a) GCN--GIN, $128$--$64$; (b) GCN--GraphSAGE, $64$--$64$; (c) GIN--GraphSAGE, $64$--$64$.}
    \label{fig:combined_analysis}
\end{figure*}

\begin{figure}[t]
    \centering
    \includegraphics[width=0.95\columnwidth]{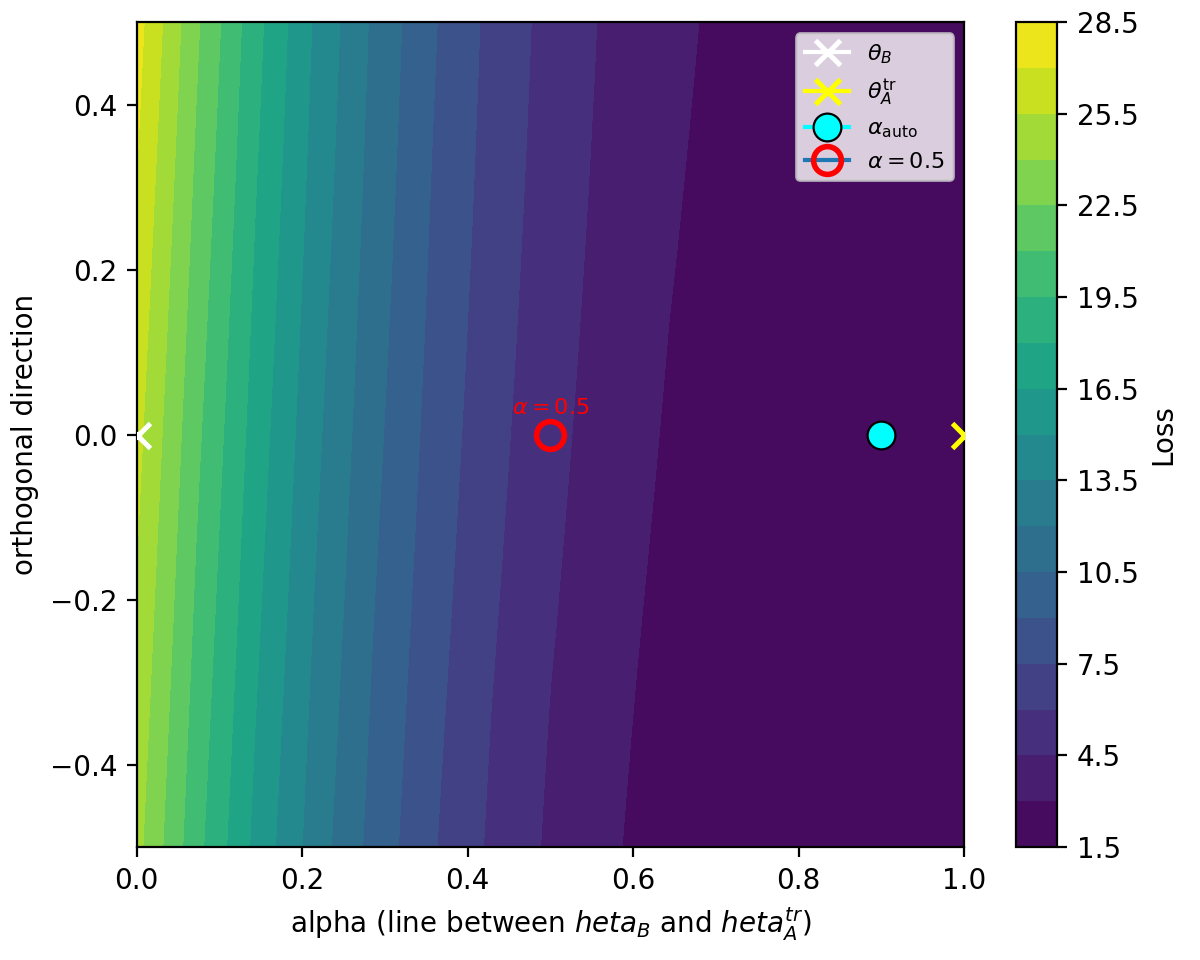}
    \caption{Loss landscape along the transport-aligned interpolation path on CiteSeer (GCN--GIN, 3--2 depth, 128--64 width). The label-free $\alpha_{\mathrm{auto}}$ (red) closely tracks the true minimum, while $\alpha=0.5$ lies far from the optimal basin.}
    \label{fig:loss_landscape}
\end{figure}

\paragraph{Label-free $\alpha$ selection versus fixed interpolation.}
A central design choice in \textsc{H-GRAMA} is the label-free confidence-risk (LF-Conf-Risk) selection of layer-wise mixing coefficients $\alpha_\ell$ (Section~\ref{sec:method:fusion}). Figure~\ref{fig:loss_landscape} visualizes the loss landscape along the transport-aligned interpolation path $\theta(\alpha) = \alpha\,\widetilde{\theta}_A + (1-\alpha)\,\theta_B$. The automatically selected $\alpha_{\mathrm{auto}}$ falls near the loss minimum, while the naive midpoint incurs substantially higher loss.

\begin{figure}[t]
    \centering
    \includegraphics[width=0.95\columnwidth]{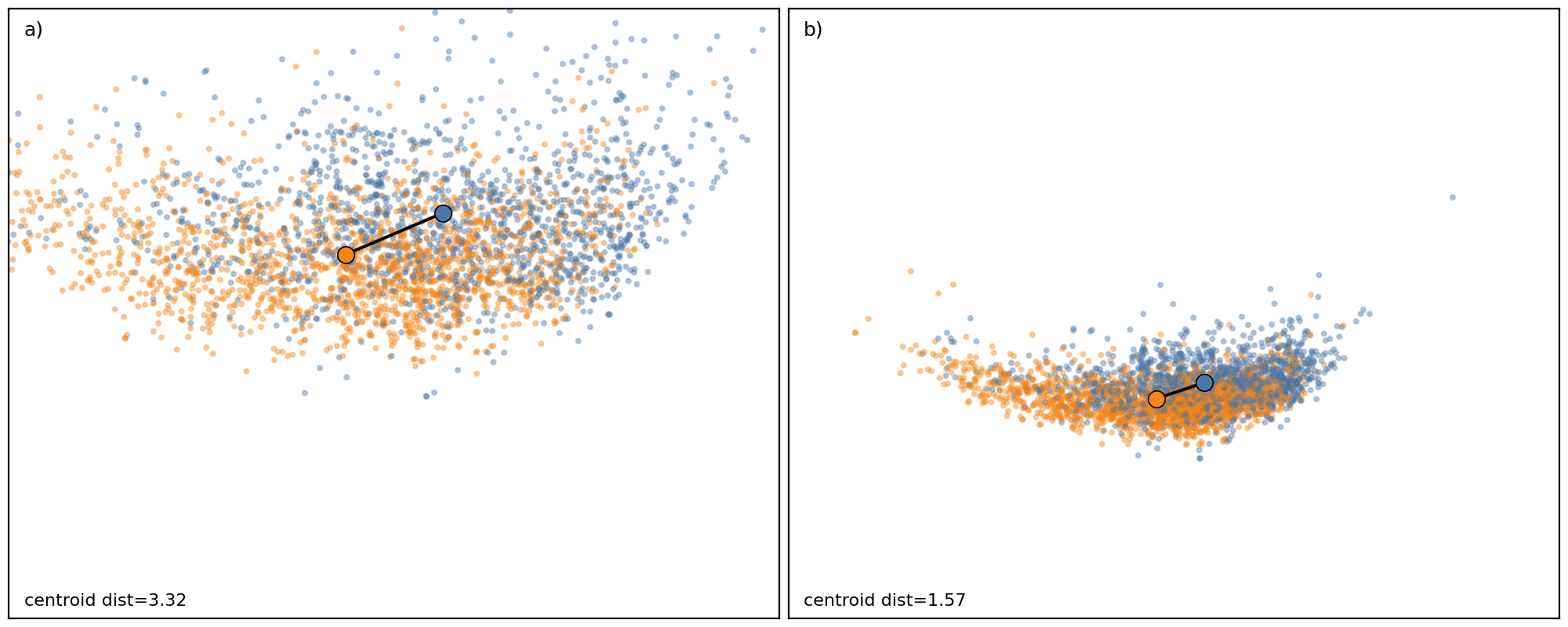}
    \caption{PCA projection of child pre-activations before (left) and after (right) LFNorm calibration on CiteSeer (GCN--SAGE, 2--2 depth, 64--64 width). Centroid distance contracts by a substantial amount, aligning the child with interpolated parent targets.}
    \label{fig:lfnorm_pca}
\end{figure}

\paragraph{LFNorm calibration effect on representation geometry.}
Figure~\ref{fig:lfnorm_pca} visualizes the pre-activation geometry (PCA projection) before and after LFNorm calibration for a representative merge. Prior to calibration, the child's class centroids are displaced from the parent targets; after LFNorm, centroids contract substantially, aligning the child's internal representation geometry with the interpolated parent target. This geometric correction translates directly to improved downstream accuracy.

\subsection{Qualitative Analysis: Learned Operator Mixtures}
\label{sec:qualitative}

To probe how \textsc{H-GRAMA} reconciles heterogeneous parent architectures at the operator level, we inspect the learned per-layer basis gate magnitudes in the merged child. Figure~\ref{fig:combined_analysis} summarizes the resulting mixtures and exposes three consistent, interpretable behaviors that support the design of UMPM.

\section{Conclusion}
We introduced \textsc{H-GRAMA}, a training-free and label-free framework for merging \emph{heterogeneous} GNN specialists trained on disjoint labeled node subsets of the same graph. The key idea is to lift merging from parameter space to a shared \emph{operator space} via Universal Message Passing Mixtures (UMPM), then align heterogeneous depths using CKA-based monotone matching, transport parameters with (semi-)orthogonal Procrustes maps, and fuse operators using closed-form gate regression and convex mixing with label-free coefficients. Finally, we mitigate merge-induced message distribution shift through LFNorm moment calibration with a memory-efficient folded implementation. Empirically, \textsc{H-GRAMA} enables cross-architecture merges (e.g., GCN$\leftrightarrow$GAT, SAGE$\leftrightarrow$GIN) that preserve specialist retention while providing meaningful inference speedups over two-model ensembles.

\paragraph{Limitations and future work.}
Performance degrades most under severe depth mismatch and in merges involving strongly MLP-dominant updates (e.g., GIN), where linear transport cannot fully capture nonlinear reparameterizations. Future work includes richer transport families beyond orthogonal maps, improved handling of MLP-heavy blocks, and extending the operator basis to cover additional message functions (e.g., edge features, positional encodings) while preserving merge-time closed-form computation.


\bibliography{references2}
\bibliographystyle{icml2026}


\end{document}